\documentclass[letterpaper, 10 pt, conference]{ieeeconf}  

\IEEEoverridecommandlockouts                              

\usepackage{amsfonts}
\usepackage{amsmath}
\usepackage{graphicx} 
\usepackage{float}
\usepackage{afterpage}
\usepackage[linesnumbered,ruled,vlined]{algorithm2e}
\usepackage{multirow}
\usepackage[table,xcdraw]{xcolor}
\usepackage{cite}
\usepackage{subfigure}
\usepackage{hyperref} 
\usepackage{pifont}

\title{\LARGE \bf
CAFE-AD: Cross-Scenario Adaptive Feature Enhancement for Trajectory Planning in Autonomous Driving
}

\author{
    Junrui Zhang$^{1,\dagger}$, Chenjie Wang$^{2,\dagger}$, Jie Peng$^{3}$, Haoyu Li$^{4}$, Jianmin Ji$^{2,4,*}$, Yu Zhang$^{2,4}$ and Yanyong Zhang$^{2,3}$
    \thanks{$^{1}$ Institute of Advanced Technology, University of Science and Technology of China (USTC), Hefei 230026, China. {\tt\small zhangjunrui@mail.ustc.edu.cn}}%
    \thanks{$^{2}$ Institute of Artificial Intelligence, Hefei Comprehensive National Science Center, 230088, China. {\tt\small wangchenjie@iai.ustc.edu.cn}}%
    \thanks{$^{3}$ School of Artificial Intelligence and Data Science, USTC, Hefei 230026, China. {\tt\small pengjieb@mail.ustc.edu.cn, yanyongz@ustc.edu.cn}}%
    \thanks{$^{4}$ School of Computer Science and Technology, USTC, Hefei 230026, China. {\tt\small lihaoyu31801@mail.ustc.edu.cn, yuzhang @ustc.edu.cn}}%
    \thanks{$\dagger$ Equal contribution to this paper.}%
    \thanks{$*$ Corresponding author. {\tt\small jianmin@ustc.edu.cn}}%
}

\begin{document}

\maketitle
\thispagestyle{empty}
\pagestyle{empty}


\begin{abstract}
Imitation learning based
planning tasks on the nuPlan dataset 
have gained
great interest
due to their potential to generate human-like driving behaviors.
However, open-loop training on the nuPlan dataset tends to cause causal confusion during closed-loop testing, and the dataset also presents a long-tail distribution of scenarios.
These issues introduce challenges for imitation learning.
To tackle these problems, we introduce CAFE-AD, a \underline{C}ross-Scenario \underline{A}daptive \underline{F}eature \underline{E}nhancement for Trajectory Planning in \underline{A}utonomous \underline{D}riving method, designed to enhance feature representation across various scenario types. We develop an adaptive feature pruning module that ranks feature importance to capture the most relevant information while reducing the interference of noisy information during training. Moreover, we propose a cross-scenario feature interpolation module that enhances scenario information to introduce diversity, enabling the network to alleviate over-fitting in dominant scenarios.
We evaluate our method CAFE-AD, on the challenging public nuPlan Test14-Hard closed-loop simulation benchmark. The results demonstrate that CAFE-AD outperforms state-of-the-art methods including rule-based and hybrid planners, and exhibits the potential in mitigating the impact of long-tail distribution within the dataset. Additionally, we further validate its effectiveness in real-world environments. The code and models will be made available at \textcolor{magenta}{\url{https://github.com/AlniyatRui/CAFE-AD}}.

\end{abstract}

\section{Introduction}
With the rapid development of artificial intelligence and autonomous driving, imitation-based motion planners are increasingly being trained to generate driving trajectories that closely mimic human behavior using observations of dynamic driving environments \cite{bansal2018chauffeurnet,zhou2021exploring,vitelli2022safetynet,scheel2022urban,cheng2024rethinking,renz2022plant,cheng2024pluto, huang2024dtpp,huang2023gameformer,guo2023ccil}. 
Many studies focus on the research of planning networks using real-world datasets\cite{caesar2020nuscenes, caesar2021nuplan}. The nuPlan dataset\cite{caesar2021nuplan} is used commonly due to its inclusion of large amounts of real driving data, long temporal trajectories, and support for both open-loop and closed-loop simulation. However, the quality of trajectories generated by directly imitating offline collected trajectories in the nuPlan dataset is often limited by the following two aspects: ~\ding{172} Causal Confusion and ~\ding{173} Long-tail Distribution.

\begin{figure}[t]
	\subfigure[] {
		\label{fig:INP-2}   
		\includegraphics[width=0.42\linewidth]{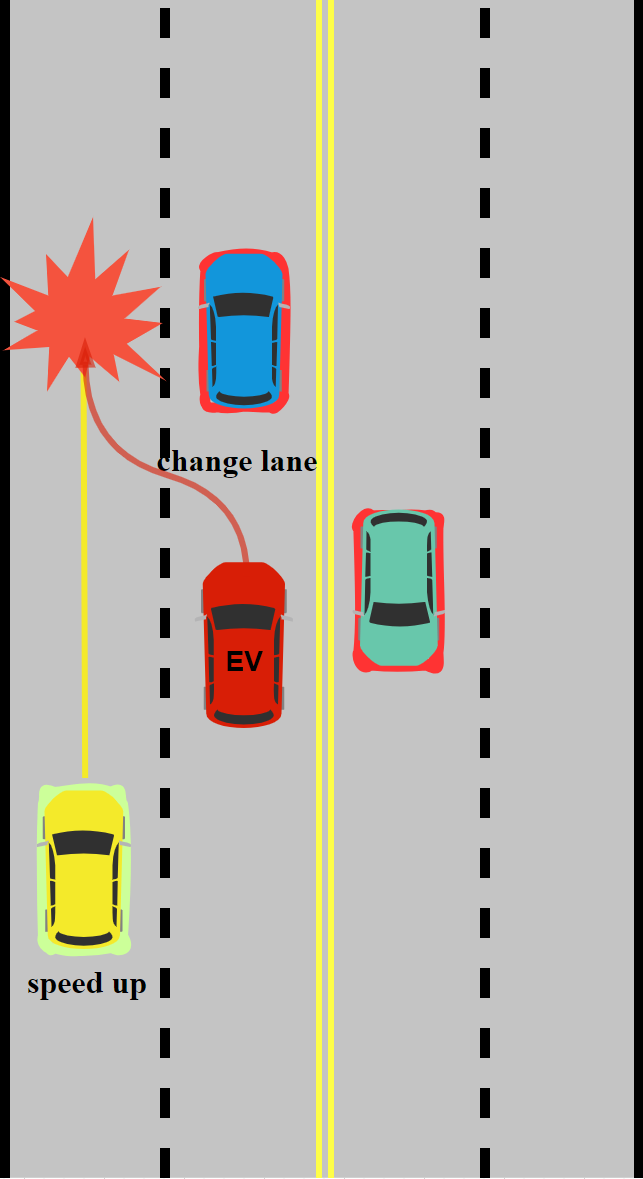}  
	} 
    \hspace{-0.4cm}
	\subfigure[] {
        \label{fig:dataset_count}  
		\includegraphics[width=0.56\linewidth]{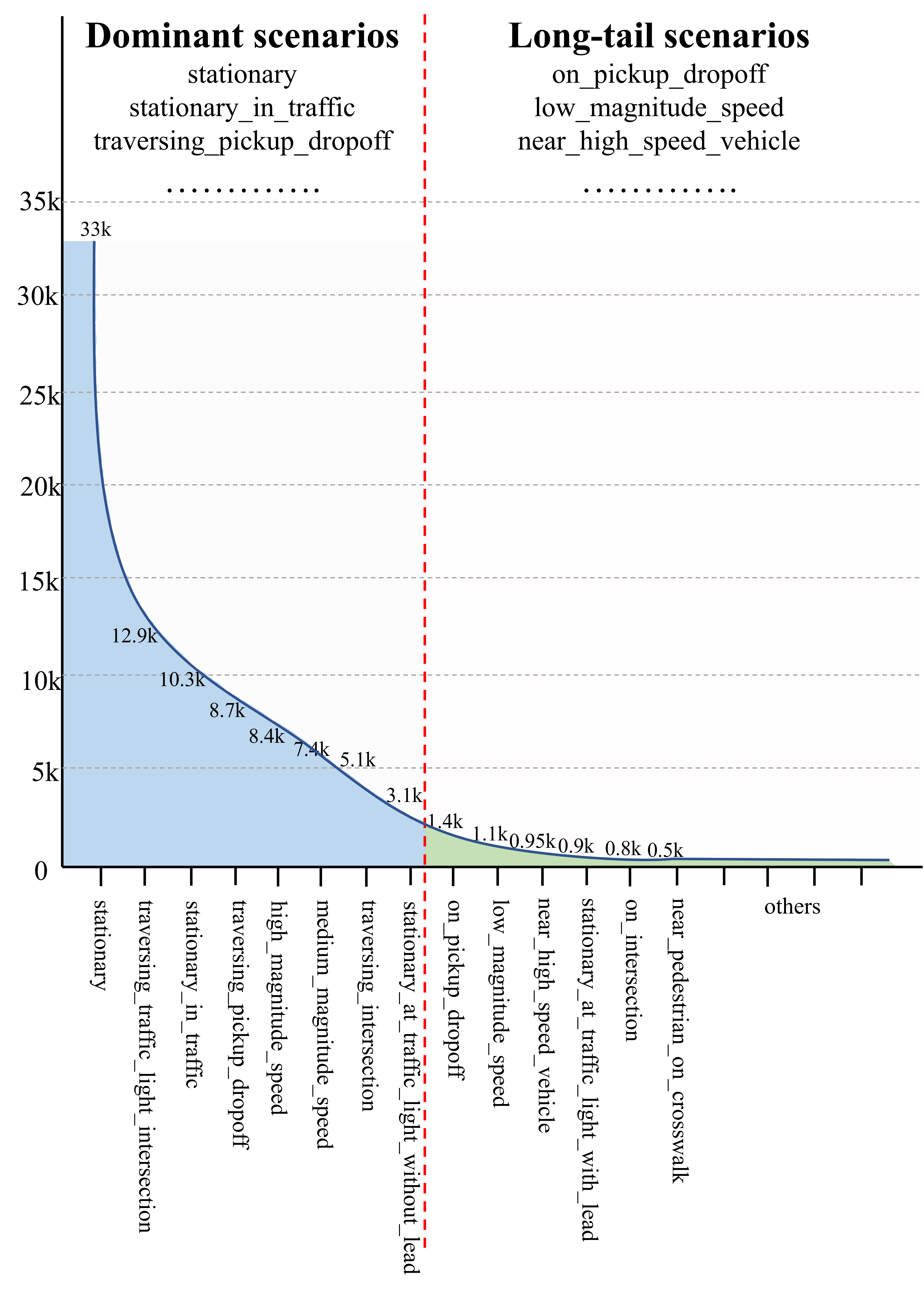}  
	} 
     \vspace{-0.4cm}
	\caption{We illustrate the inherent problems in the dataset. (a) Causal Confusion: Despite the green vehicle being closest to the ego vehicle, it has no causal relationship with the ego's planning. However, the ego vehicle may overly focus on it and neglect the yellow vehicle behind, leading to a potential collision. (b) Long-tail Distribution: Due to the limitations in data collection caused by the complexity of autonomous driving, the distribution across different scenarios is highly imbalanced, which easily leads to the model over-fitting to the dominant scenario, e.g. $stationary$.}
	\label{fig:INP}
    \vspace{-0.65cm}
\end{figure}

\paragraph{Causal Confusion} 
The open-loop training method in nuPlan, which relies on pre-collected trajectories for supervision, tends to cause the model to overly depend on the results of expert behavior. As a result, the model struggles to distinguish causal relationships in the environment, inadvertently focusing on unexpected noise rather than the critical information in the current scenario, as illustrated in Fig. \ref{fig:INP-2}. 

PLUTO\cite{cheng2024pluto} uses hand-crafted modules to remove non-reactive agents from future ground truths. It relies on predefined rules, making it less adaptable to dynamic and complex driving environments. Inspired by previous works\cite{rao2023dynamic}, \cite{kong2023peeling}, which employ dynamic token sparsification to allow the network to focus effectively on most informative regions. We aim to further address the issue of causal confusion by grasping the most critical information for the planning process.
Specifically, we propose an Adaptive Pruning Transformer Encoder that adaptively identifies the importance of different tokens (e.g., agents, obstacles) using attention probabilities during the encoding phase. It removes tokens and attention connections deemed unimportant by the model, effectively filtering out irrelevant information.

\paragraph{Long-tail Distribution}
There is a significant disparity in the scale of different scenario types within the dataset, as illustrated in Fig. \ref{fig:dataset_count}. Dominant scenarios tend to cause the model to overfit their data distribution, leading to poor performance in long-tail scenarios. 
Previous research\cite{liu2024cross,yao2022improving} improves generalization in model training through mix-up in the feature space.
Inspired by these methods, we propose the cross-scenario feature interpolation to address the over-fitting to dominant scenarios caused by the long-tail distribution in the dataset. By introducing a scenario classifier to decompose features into scenario-relevant and scenario-generic components, we perform feature enhancement through the interpolation of scenario-relevant features across different scenario types. Thereby, this module improves the model's generalization ability and robustness across diverse scenarios.

Our proposed CAFE-AD (Fig. \ref{fig:system_overview}), Cross-Scenario Adaptive Feature Enhancement for Trajectory Planning in Autonomous Driving, effectively mitigates the interference of the aforementioned challenge, enhancing the planner's ability to capture critical scene information and improving the representation capability across various scenarios. 
Leveraging our approach, CAFE-AD demonstrates competitive performance compared to current state-of-the-art methods in both closed-loop interactive and non-interactive simulations on the Test14-Hard Benchmark. Additionally, we deploy our designed planner in real-world scenarios on self-driving platforms to generate reference trajectories and further validate its effectiveness.

The primary contributions of our research are outlined as follows:
\begin{itemize}
    \item We propose CAFE-AD, an effective feature enhancement method designed to improve the performance of imitation learning based planners, ensuring safer and more efficient autonomous driving.
    \item The design of the Adaptive Pruning Transformer Encoder specifically enhances the model’s ability to focus on critical scene elements by selectively pruning irrelevant information. It improves the model's understanding of the causal relationships in the current driving scenario, leading to better performance.
    \item The cross-scenario feature interpolation is proposed to enhance the information in dominant scenarios with scenario-relevant features from different types. It improves the diversity of scenarios during training, preventing the over-fitting caused by the long-tail distribution.
    \item Our method demonstrates competitive performance compared to state-of-the-art methods on the Test14-Hard Benchmark. Furthermore, we validated the effectiveness of our approach in real-world scenarios, underscoring its practical value.
    
\end{itemize}

\section{Related Work}

\subsection{Imitation Learning}
Imitation learning based motion planners are primarily divided into two categories: end-to-end (E2E) approaches\cite{codevilla2019exploring,chen2020learning, chitta2022transfuser, vitelli2022safetynet, ye2023fusionad, jia2023think}, which directly generate driving trajectories or control signals from raw sensor inputs, and mid-to-mid approaches\cite{huang2023gameformer, hu2023imitation, xi2023imitation} which utilize processed perception outputs as inputs.

Some E2E approaches \cite{hu2023planning,chen2024vadv2,jiang2023vad,chen2022learning} have integrated perception, prediction, and planning into a unified architecture, where each module produces its own output while sharing extracted features with downstream tasks to form an end-to-end output. With the advancement of large language models (LLMs), the recent work\cite{xu2024drivegpt4,wang2023drivemlm,shao2024lmdrive} has started to explore the integration of LLMs to leverage their scene understanding and logical reasoning capabilities. 
However, E2E methods need large amounts of quality data and are hard to debug due to their lack of interpretability, making real-world applications challenging despite validation in simulators like CARLA \cite{dosovitskiy2017carla}.

This paper primarily focuses on mid-to-mid approaches, which are more modular and interpretable, making them widely applicable to real-world data\cite{bansal2018chauffeurnet,shao2023safety,scheel2022urban}. 
The release of the nuPlan dataset \cite{caesar2021nuplan} has provided recorded real-world training data and a simulator for both open-loop and closed-loop testing. The PDM method\cite{dauner2023parting}, which relies solely on the centerline rather than more complex scene representations \cite{renz2022plant, scheel2022urban, hanselmann2022king, pini2023safe, hallgarten2023prediction} 
achieved first place in the nuPlan challenge 2023. In the same competition, several methods\cite{huang2023gameformer, hu2023imitation, xi2023imitation} 
achieved high scores by combining motion planning networks with post-processing optimization techniques. 
Additionally, other intriguing works include integrating occupancy prediction to guide planning\cite{liu2023occupancy}, leveraging a Multi-modal LLM to integrate human-like reasoning and common sense\cite{zheng2024planagent}, the tree-structured policy planning with differentiable joint training for motion prediction and cost evaluation\cite{huang2024dtpp}, endeavoring to mask the ego vehicle's information\cite{guo2023ccil}.


However, pure imitation learning typically suffers from limitations such as causal confusion and long-tail distribution, which result in poor performance of models. 
To address this, our approach tackles causal confusion by selecting more informative features from the current driving scenario and mitigates the long-tail distribution problem by introducing diverse representations from other scenario types.

\begin{figure*}[t]
\centering
\includegraphics[width=\linewidth]{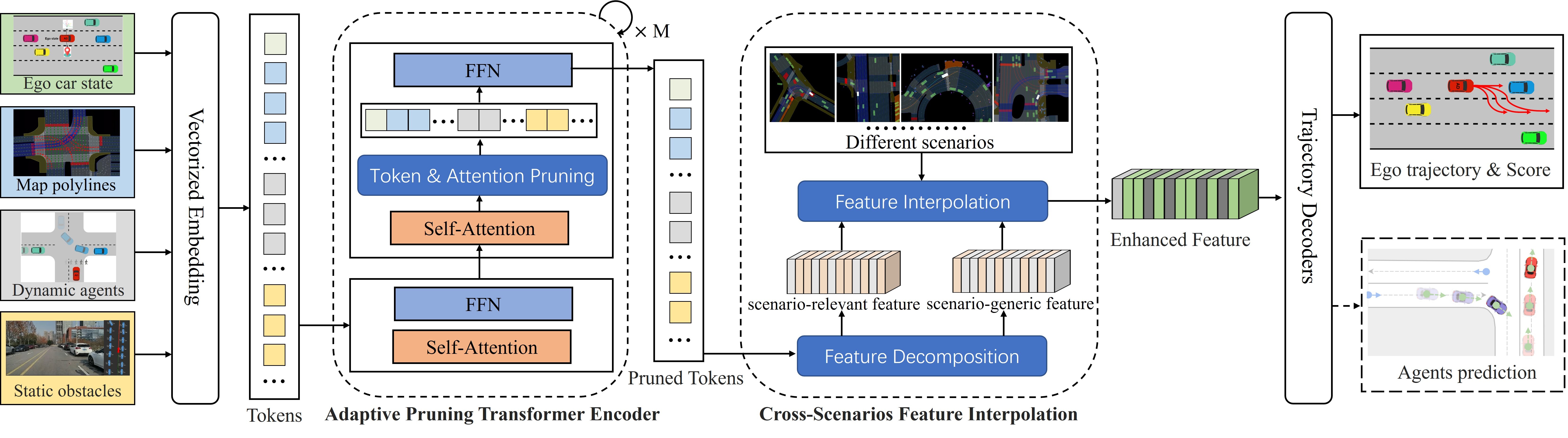}
\vspace{-0.7cm}
\caption{The overview of our proposed CAFE-AD method. 
}
\label{fig:system_overview}
\vspace{-0.6cm}
\end{figure*}

\subsection{Enhancement Strategies}
Several studies have aimed to overcome the limitations of pure imitation learning by employing augmentation techniques that enhance the learning process, thereby improving the models' applicability and effectiveness in diverse and complex scenarios. The most prevalent enhancement strategies commence at the input data level, employing techniques such as data perturbation\cite{bansal2018chauffeurnet,cheng2024rethinking} and reasonable trajectory augmentation\cite{xi2023imitation} to introduce variability into the learning process. PLUTO\cite{cheng2024pluto}, PlanTF\cite{cheng2024rethinking} address causal confusion during training by adding a Dropout layer to input features. Additionally, incorporating supplementary loss functions, for instance, those that penalize collisions and deviations from the designated path \cite{bansal2018chauffeurnet,zhou2021exploring,scheel2022urban} is also a critical strategy for enhancing the model's overall efficacy.

Compared to these methods that typically perform enhancement in the input space or add auxiliary loss function, our research introduces a novel attempt to achieve more versatile performance through cross-scenario adaptive feature enhancement in the encoded space.

\section{Methodology}
In section \ref{sec:scenerepresentation}, we first introduce the model's scene representation input. We then delve into the specifics of the feature enhancement framework, covering the adaptive pruning module in Section \ref{sec:adaptivepruning} and the cross-scenario feature interpolation module in Section \ref{sec:crossscenario}. The comprehensive workflow of our research is illustrated in Fig. \ref{fig:system_overview}.

\subsection{Scene Representation} 
\label{sec:scenerepresentation}
In this study, we focus on autonomous driving trajectory planning in urban traffic environments, which involves key elements such as the ego vehicle ($EV$), surrounding dynamic agents ($A$), high-definition maps ($M$), static obstacles ($O$), and various traffic semantic information. We use PLUTO as the baseline for our method, initially encoding information from different traffic elements into a common feature dimension $D$, where the encoded vectors are denoted as $X_{EV}, X_{A}, X_{M}, X_{O}$. To achieve information sharing, we concatenate different vectors into a tensor. To compensate for the loss of global positional information during vectorization, each token is introduced with a Fourier positional embedding $X_{PE}$, and traffic semantic information is also encoded as learnable embedding $X_{LE}$ and added to the tensor, forming the initial scene understanding embedding $X_0$ as \begin{equation}\label{eq1} X_0 = \text{concat}(X_{EV}, X_{A}, X_{M}, X_{O}) + X_{PE} + X_{LE} \end{equation} where the $X_0$ as input then processed by a multi-layer Adaptive Pruning Transformer Encoder denoted as APT-Encoder, where tokens corresponding to different traffic elements are reordered based on their adaptive importance ranking.

\subsection{Adaptive Pruning Transformer Encoder}
\label{sec:adaptivepruning}
Causal confusion arises because the model struggles to retrieve crucial features from the complex driving environment. To address this, we developed an adaptive pruning transformer encoder, applying token- and attention-level pruning to help the model focus on features that are more important for trajectory planning.
After being fed into the APT-Encoder, 
$X_0$ is initialized using self-attention\cite{vaswani2017attention}, as follows. The processed results are then divided into features corresponding to each traffic element in Eq.~\eqref{eq1}.
\begin{equation}
\text{Self-Attention}(X_0) = [F_{EV}, F_{A}, F_{M}, F_{O}]
\end{equation}
where, in the self-attention module, the input is passed through three linear transformations to generate the query ($Q$), key ($K$), and value ($V$) matrices. The self-attention matrix $A$ is calculated as follows.
 \begin{equation} A =  \text{Softmax}( QK^T/\sqrt{D})V \end{equation}

Here we denote the token length of $X_0$ as $L$, the query matrix $Q$ and the key matrix $K$ are constructed by concatenating the token representations, where $Q = [q_1,q_2,q_3...q_L]$ and $K = [k_1,k_2,k_3...k_L]$. We use the attention map of the ego-vehicle state token computed using the following equation, to represent the contribution of other tokens to the ego-vehicle's planning performance. 
 \begin{equation} 
 a_{EV} = \text{Softmax}(q_{EV}K^T/\sqrt{D})
 \end{equation}
 
We utilize $\tilde{a}_{EV}$ after excluding the first element of $a_{EV}$, as the criterion for selecting tokens. In the multi-head self-attention mechanism, the average attention probabilities $\bar{a}_{EV} = \frac{1}{H} \sum_{i=1}^{H} \tilde{a}_{EV}^{(h)}$ across all heads are used as the basis for ranking and selecting tokens. Therefore, for token pruning, based on the values of $\bar{a}_{EV}$, we select a predefined number of the most important tokens from $F_{A}, F_{M} \text{ and } F_{O}$ according to a specified ratio $\pi_{p}$. These selected tokens are then concatenated to obtain pruned feature $F$ (Eq.~\eqref{eq2}). In attention-level pruning, we select specific attention connections. Specifically, each token computes attention only with the selected tokens.
 \begin{equation}\label{eq2} 
 F = \text{concat}(F_{EV}, \text{select}(F_{A}), \text{select}(F_{M}), \text{select}(F_{O}))
 \end{equation}

After processing by the APT-Encoder, the pruned feature with more relevant information is obtained and fed into the Cross-Scenario Feature Interpolation module denoted as CSFI for diversity enhancement.

\subsection{Cross-Scenario Feature Interpolation} 
\label{sec:crossscenario}
While the pruning mechanism helps the model better capture crucial intra-scenario information, it does not address the optimization challenge caused by the long-tail distribution of the overall dataset. To mitigate the model over-fitting to the dominant scenario, we developed a cross-scenario feature interpolation method. We initially use a scenario classifier to predict the scenario category based on the encoded features $F$. The cross-entropy loss $\mathcal{L}_{\text{CE}}$ is calculated between the predicted scenario category and the ground truth. The gradient of the loss with respect to the encoded features $F$ is denoted as $\nabla_F \mathcal{L}_{\text{CE}}$. We compute the contribution of each feature dimension $i$ to the scenario classifier's predicted logit $C_i$ as: 
\begin{equation} 
C_i = \text{Mean}(F_i \cdot \nabla_{F_i} \mathcal{L}_{CE}, dim = 0)
\end{equation} 

Then we determine the threshold $\tau$ based on the ratio $\pi_o$ by calculating the quantile of the contributions.
\begin{equation}
\tau = \text{Quantile}( [C_0, C_1,\dots, C_D] , \pi_o)
\end{equation}
where the threshold $\tau$ is used to decompose the features along the embedding dimension into two components, as shown below:
\begin{equation}
F = \mathbb{I}\left({C}_i > \tau \right) \odot F + \mathbb{I}\left({C}_i \leq \tau \right) \odot F
\end{equation}
where the former represents features with higher relevance to specific scenarios, denoted as scenario-relevant features $F_r$, while the latter represents scenario-generic features $F_g$ that are shared across different scenarios. Here, \(\odot\) denotes the element-wise multiplication.

We enhance the scenario-relevant feature $F_r$ for introducing further diversity through interpolation with scenario-relevant features $F_r'$ from different scenarios. Here $\pi_r$ represents the interpolation ratio, which is sampled from a uniform distribution $\mathcal{U}(0, 1)$. Note that only the dominant scenario is augmented. In our experiments, dominant scenarios are defined as those whose quantities exceed the average number per scenario. The process is shown below:
\begin{equation}F_{interpolation} = F_g + (1-\pi_r) \times F_r + \pi_r \times F_r'\end{equation}

Finally, the enhanced features are passed through a Transformer decoder to generate the ego vehicle's future trajectory and predict the behavior of dynamic agents. By incorporating features from different scenarios, we introduce diversity to encourage the model to balance fitting across scenarios. 

\subsection{Training Process} 
We denote $\phi$ as the feature decoder. The loss function takes the feature $E$ and the decoder's output $\phi(E)$ as inputs, following the design of PLUTO\cite{cheng2024pluto}, which incorporates contrastive learning loss, auxiliary loss, ego-trajectory imitation loss, and surrounding agents' prediction loss. 
\begin{equation} \label{equ10}
    \mathcal{L}(E, \phi(E)) = \mathcal{L}_{cll} + \mathcal{L}_{aux} + \mathcal{L}_{ego} + \mathcal{L}_{agents}
\end{equation}

We train the model in two phases. During the warm-up phase, the planning model is trained on the original encoded feature $E$ following $\mathcal{L}(E, \phi(E))$. The scenario category classifier $\theta$ is trained using $E$ to predict the scenario category labels $i$ with $\mathcal{L}_{CE}(\theta(E), i)$.

After the warm-up phase is completed, we use both the original and augmented features with the following objective function:
\begin{equation} \label{equ11}
    \mathcal{L}_{{aug}} = \frac{1}{2} \left( \mathcal{L}(E, \phi(E)) + \mathcal{L}(E', \phi(E')) \right)
\end{equation}
where the $E'$ represents the augmented feature from the original feature $E$, notice that due to the pruning process, the agents are reordered, thus the prediction ground truth for $\mathcal{L}_{agents}$ in Equation \ref{equ10} is modified correspondingly. 
Meanwhile, the scenario category classifier uses the pruned token to compute the loss with $\mathcal{L}_{CE}$ and updates accordingly, where the token does not use interpolation to change the distribution but simply drops a few insignificant features.

\begin{table*}[]
\centering
\caption{Comparison with the state-of-the-art methods.}
\vspace{-0.2cm}
\label{table:comparison}
\renewcommand{\arraystretch}{1.25}
\begin{tabular}{ll|ccccccc|c}
\hline
Type                           & Methods    & R-Score & Collisions & TTC & Drivable & Comfort & Progress & Speed & NR-Score \\ \hline
Expert                         & Log-Replay & 68.80   & 77.02      & 69.85 & 95.96    & 99.26   & 98.48    & 94.12 & 85.96    \\ \hline
\multirow{2}{*}{Rule-based}    & IDM        & 62.26   & 84.38      & 72.43 & 84.19    & 87.87   & 69.6    & 96.52 & 56.16    \\
                               & PDM-Closed & 75.19   & 95.22      & 84.19 & 95.59    & 83.46   & 75.48    & \textbf{99.54} & 65.08    \\ \hline
\multirow{5}{*}{Pure-learning} & UrbanDriver      & 36.39   & 63.05      & 56.62 & 72.43    & \textbf{99.63}   & \textbf{79.99}    & 79.19 & 35.71    \\
                               & DTPP       & 42.67   & 88.6      & 84.93 & 94.12    & 94.12   & 38.26    & 99.28 & 41.60    \\
                               & PlanTF     & 58.76   & 89.52      & 83.82 & 91.91    & 83.82   & 65.38    & 97.79 & 60.53    \\
                               & PLUTO (w/o post.)      & 58.74   & 87.68      & 81.62 & 94.12    & 86.03   & 64.87    & 97.71 & 63.24    \\
                               & CAFE-AD (w/o post.)       & 65.25   & 88.6      & 82.35 & 94.85    & 88.6   & 71.64    & 97.43 & 68.84    \\ \hline
\multirow{3}{*}{Hybrid}        & GameFormer & 69.09   & 92.10      & 82.72 & 91.91  & 91.91  & 68.73   & 98.00     & 66.74   \\
                               & PLUTO      & 75.35   & 95.59      & 87.87 & 97.06    & 78.68   & 75.07    & 97.8 & 73.87    \\
                               & CAFE-AD (Ours)       & \textbf{78.16}   & \textbf{95.96}      & \textbf{88.97} & \textbf{97.43}    & 83.82   & 77.51    & 97.93 & \textbf{76.04}    \\ \hline
\end{tabular}
\vspace{-0.5cm}
\end{table*}

\section{Experimental Setup}
\subsection{Implementation Details} 
In this work, we propose an augmentation method to enhance the driving performance of imitation learning planners. Our prune ratio $\pi_{p}$ is set to 0.9 and is applied every two layers of the Transformer Encoder. 
We implement a cyclic threshold adjustment scheme for the scenario-relevant threshold $\pi_{o}$ that enables the model to adapt incrementally to increasing enhancement strength. The $\pi_{o}$ is initiated with a threshold of 0.9, which is decremented by 0.1 every 100 steps, reaching a minimum of 0.5 before resetting to 0.9 to repeat the process. In accordance with the nuPlan challenge 2023 design, we use 2 seconds of historical data to predict 8 seconds of future trajectory. The model is trained on 4 NVIDIA A30 GPUs with a batch size of 128 for 25 epochs. We use the AdamW optimizer with a weight decay set to 1$e^{-4}$. The learning rate is linearly increased to 1$e^{-3}$ over the initial three warm-up epochs, followed by a cosine decay schedule for the rest of the training. 

\subsection{Evaluation Setting} 
For all experiments, we use a consistent training split of 100k frames sampled from all scenario types for training. For evaluation, we use the nuPlan Test14-Hard Benchmark which is challenging for imitation learning based planners. As detailed in PlanTF\cite{cheng2024rethinking}, the Test14-Hard Benchmark was created by running 100 scenarios for each of the 14 scenario types specified in the nuPlan challenge 2023 using a powerful rule-based planner (PDM-Closed) and then selecting the 20 least-performing scenarios for each type. 
The nuPlan dataset consists of three simulation challenges: open-loop simulation, non-reactive closed-loop simulation, and reactive closed-loop simulation. Previous work \cite{phong2024truly} has demonstrated a significant gap between the open-loop evaluation metrics and real driving performance, so we only conduct closed-loop simulations. The evaluation metrics for closed-loop simulations include: (1) Traffic Rule Compliance: The evaluation of the EV's adherence to speed limits, maintenance of the correct driving direction, and the ability to stay within the drivable area boundaries. (2) Safety: The assessment of vehicle safety by checking the avoidance of ego-at-fault collisions and ensuring that the Time to Collision (TTC) exceeds a specified threshold. (3) The evaluation of comfort through acceleration levels and the assessment of progress by comparing route coverage to that of the expert driver.
\subsection{SOTA Methods}
We compare current state-of-the-art methods with our model. (1) \textbf{IDM}\cite{treiber2000congested}: A classic rule-based method that adjusts vehicle behavior. (2) \textbf{PDM-Closed}\cite{{dauner2023parting}}: An advanced rule-based algorithm that won the nuPlan challenge 2023. It generates multiple candidate trajectories through various hyper-parameters of IDM and ultimately selects the optimal trajectory. (3) \textbf{UrbanDriver}\cite{scheel2022urban}: A planner that uses vectorized inputs with polyline encoders based on PointNet and Transformers. Its open-loop variant is employed for evaluation, incorporating history perturbation during training. (4) \textbf{GameFormer}\cite{huang2023gameformer}: A k-level game-based interactive trajectory prediction and motion planning architecture. (5) \textbf{DTPP}\cite{huang2024dtpp}: An integrating joint training method for ego-conditioned motion prediction and cost evaluation within a tree policy planner. 
(6) \textbf{PlanTF}\cite{cheng2024rethinking}: A concise and efficient learning-based baseline based on the Transformer architecture. 
(7) \textbf{PLUTO}\cite{cheng2024pluto}: A powerful method based on a longitudinal-lateral aware model architecture, enhanced with auxiliary loss techniques and contrastive learning to further improve performance.

All methods are trained on our 100k-split random datasets, following the configurations provided in their open-source versions, except for DTPP due to the failure in data processing. For DTPP, we ensure that the scenarios in the training set are contained in our split, and we introduce additional data for validation. Notice that both PLUTO and CAFE-AD initially encountered simulation errors in approximately 2\% of the Test14-Hard Benchmark. These errors were due to failures in the post-processing in some frames, so we used pure learning results to prevent these errors in those frames. Our modifications will be released along with the code.

\section{Experiments}
\subsection{Quantitative Results}

The comparative analysis with current state-of-the-art methods on the Test14-hard benchmark is detailed in Table \ref{table:comparison}. Our proposed CAFE-AD significantly outperforms all other methods across both evaluation tasks. Compared to the current leading method, PLUTO, our approach achieves a 2.81 score improvement in CLS-R and a 2.17 score improvement in CLS-NR. 
Moreover, when considering the pure learning variant without post-processing (w/o post.), we further demonstrate a more significant advantage over PLUTO's variant with a lead of 6.51 and 5.6 in score on the two tasks respectively. 
Notably, we surpass PLUTO across nearly all metrics, regardless of whether post-processing is applied. In particular, we show a great improvement in the $Progress$ metric. This is primarily because we mitigate over-fitting to dominant scenarios like $stationary$, allowing the pure learning approach to exhibit more proactive driving behavior without relying on rule-based guidance.

\begin{figure*}[t]
\centering
\includegraphics[width=\linewidth]{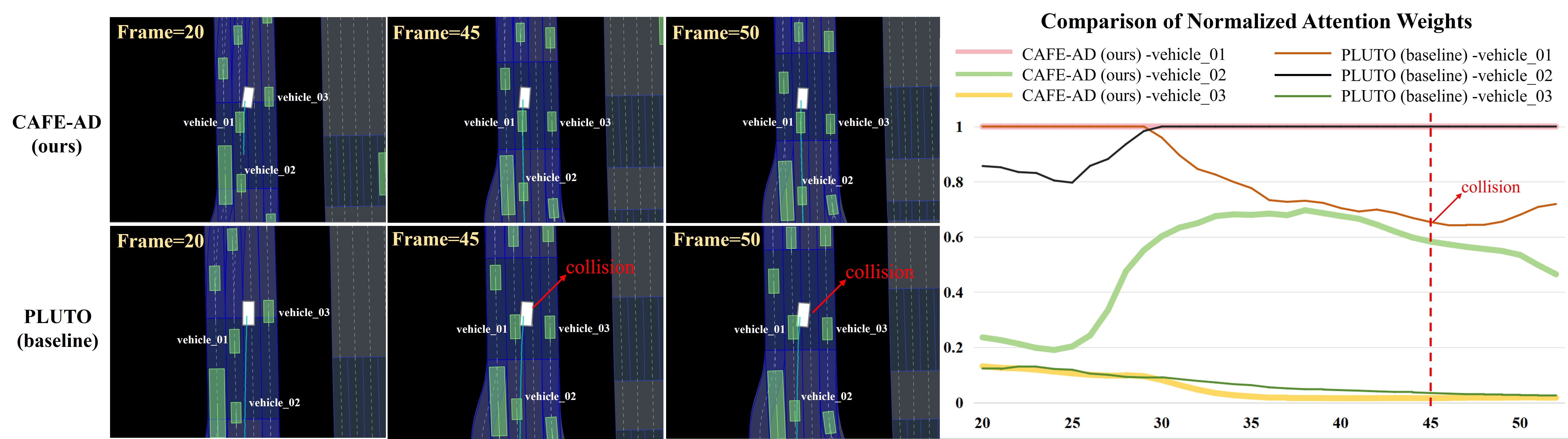}
\vspace{-0.6cm}
\caption{Qualitative evaluation between CAFE-AD (ours) and PLUTO (baseline) and the changes in normalized attention weights.}
\label{fig:attention_weight}
\vspace{-0.2cm}
\end{figure*}

\begin{figure*}[t]
\centering
\includegraphics[width=\linewidth]{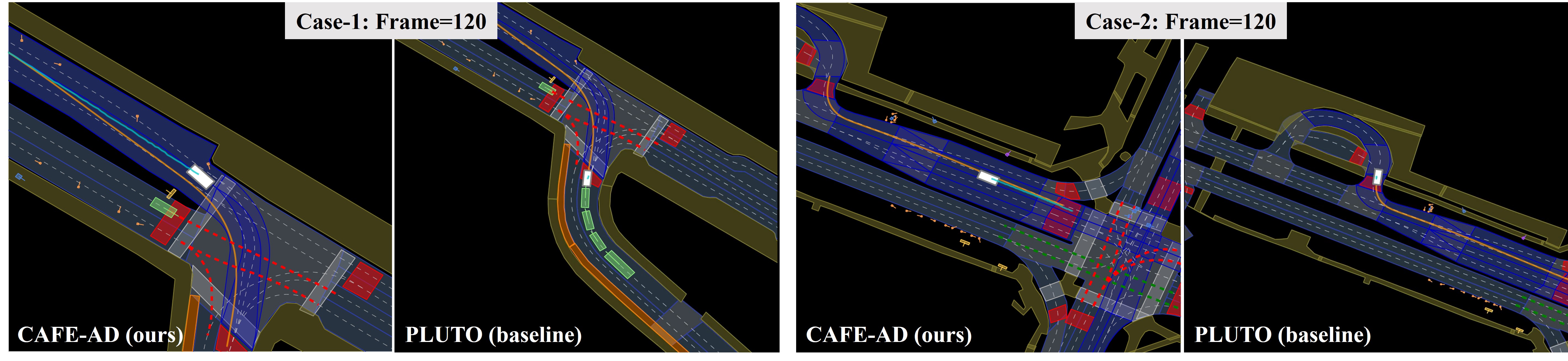}
\vspace{-0.6cm}
\caption{Qualitative evaluation of handling the long-tail distribution problem.}
\label{fig:long_tail}
\vspace{-0.3cm}
\end{figure*}

\subsection{Case Analysis}
Fig. \ref{fig:attention_weight} presents a qualitative analysis of the driving behavior between our method and the baseline PLUTO in the typical scenario from the Test14-hard benchmark. It also shows the changes in normalized attention weights toward surrounding agents at different driving moments. As shown in the figure, our method consistently identifies the $vehicle\_{01}$ ahead as the most critical agent (with a normalized attention weight of 1). In contrast, PLUTO alternates its focus between $vehicle\_{01}$ and $vehicle\_{02}$. This causes PLUTO to fail in making a timely and safe interaction with vehicle $vehicle\_{01}$, resulting in a collision at Frame 45. In comparison, our method uses the APT-Encoder to focus on the most important information in the current driving scene, reducing the impact of causal confusion. It begins deceleration at Frame 20, successfully avoiding a collision with $vehicle\_{01}$ which has the most significant impact on the current driving behavior, ensuring safe driving. 

Fig. \ref{fig:long_tail} illustrates how our method alleviates the issue of over-fitting to dominant scenarios caused by the long-tail distribution. In nuPlan dataset, approximately 45\% of the scenarios are static, e.g. $stationary$, $stationary\_in\_traffic$. As shown in the figure, this leads the PLUTO method to remain stationary in some scenarios where expert trajectories suggest moving forward, due to over-fitting to static scenarios. In contrast, our method continues to exhibit proactive driving behavior, demonstrating the effectiveness of our cross-scenario feature interpolation.

\subsection{Ablation Study}
(1) \textbf{Ablation Study of Module Design}. Our method incorporates two crucial modules: Adaptive Pruning Transformer Encoder (APT-Encoder) and Cross-Scenario Feature Interpolation (CSFI). We conducted ablation studies to validate the effectiveness of our design, as shown in Table \ref{table:ablation01}.
It demonstrates that using either the APT-Encoder or CSFI module individually results in improvements over the baseline method. The combination of these modules achieves the best overall performance. This further demonstrates that the combined enhancement from these two modules, through selecting important information within the current scenario and introducing diversity across scenarios, significantly improves planning effectiveness.

(2) \textbf{Ablation Study of Different Baseline}. Table \ref{table:ablation02} demonstrates the enhancement effect of our method on another baseline-PlanTF. The results show improvements across nearly all metrics, demonstrating the generalization and transferability of the proposed method.

\begin{table}[]
\centering
\caption{Ablation Study of Module Design.}
\vspace{-0.3cm}
\label{table:ablation01}
\renewcommand{\arraystretch}{1.26} 
\setlength{\tabcolsep}{3.5pt}
\begin{tabular}{cc|ccccc}
\hline
APT & CSFI & R-Score & Collisions & Comfort & Progress & NR-Score  \\ \hline
 & & 58.74 & 87.68 & {86.03} & 64.87 & 63.24\\ \hline
$\surd$ & & 64.24 & \textbf{91.36} & 84.56 & 64.01 & 65.2\\ 
 & $\surd$ & 64.95 & 88.79 & {86.03} & 67.0 & 66.73\\ 
$\surd$ & $\surd$ & \textbf{65.25} & {88.6} & \textbf{88.6} & \textbf{71.64} & \textbf{68.84}\\  \hline
\end{tabular}
\vspace{-0.3cm}
\end{table}

\begin{table}[]
\centering
\caption{Ablation Study of Different Baseline.}
\vspace{-0.3cm}
\label{table:ablation02}
\renewcommand{\arraystretch}{1.26} 
\setlength{\tabcolsep}{2pt}
\begin{tabular}{c|ccccc}
\hline
Method & R-Score & Collisions & Comfort & Progress & NR-Score  \\ \hline
PlanTF & {58.76} & \textbf{89.52} & 83.82 & 65.38 & 60.53\\ 
PlanTF w. CAFE-AD & \textbf{61.29} & 87.87 & \textbf{86.03} & \textbf{70.29} & \textbf{62.4}\\ \hline
\end{tabular}
\vspace{-0.3cm}
\end{table}

\subsection{Real-World Validation}
To demonstrate the effectiveness of our proposed method in real-world applications, we trained the model using expert data collected from an actual vehicle and deployed it on an autonomous driving platform. The model was integrated to provide a reference trajectory for a traditional optimization-based planning method. This hybrid method further illustrates the method's effectiveness in handling interactions with surrounding agents and enhancing safe driving. A demonstration of the real-world scenario can be seen in the attached video.

\section{Conclusions}

In this work, we propose CAFE-AD, a novel cross-scenario adaptive feature enhancement method for trajectory planning in autonomous driving, designed to address causal confusion and mitigate the long-tail distribution issues during the imitation learning training process. CAFE-AD introduces an adaptive pruning transformer encoder that filters out planning-irrelevant information, encouraging the model to focus on critical features. Furthermore, CAFE-AD employs a cross-scenario feature interpolation module to prevent the model from over-fitting to dominant scenarios in the long-tail distribution dataset. Owing to these enhancements, our method achieves competitive and generalizable results on the nuPlan Test14-Hard benchmark and improves the overall driving performance. Additionally, both quantitative and qualitative results demonstrate the effectiveness of the two proposed modules in addressing complex and dynamic driving environments.







\bibliographystyle{IEEEtran}
\bibliography{IEEEabrv,Ref}  

\end{document}